# Exploration of COVID-19 Discourse on Twitter: American Politician Edition

Cindy Kim, Daniela Puchall, Jiangyi Liang, Jiwon Kim
New York University

**Abstract**

The advent of the COVID-19 pandemic has undoubtedly affected the political scene worldwide and the introduction of new terminology and public opinions regarding the virus has further polarized partisan stances. Using a collection of tweets gathered from leading American political figures online (Republican and Democratic), we explored the partisan differences in approach, response, and attitude towards handling the international crisis. Implementation of the bag-of-words, bigram, and TF-IDF models was used to identify and analyze keywords, topics, and overall sentiments from each party. Results suggest that Democrats are more concerned with the casualties of the pandemic, and give more medical precautions and recommendations to the public whereas Republicans are more invested in political responsibilities such as keeping the public updated through media and carefully watching the progress of the virus. We propose a systematic approach to predict and distinguish a tweet's political stance (left or right leaning) based on its COVID-19 related terms using different classification algorithms on different language models.

**1 Introduction**

In a time of crisis, as in the time of a pandemic, world leaders hold the responsibility to communicate with the people that they represent in a variety of ways. Unlike any time before, social media now plays a large role in how information is distributed. In March 2020, the World Health Organization declared the novel Coronavirus 2019[1] a global pandemic due to the high levels of spread and severity of the virus. As of December 2020, there are a reported 71.4 million cases and 1.6 million deaths worldwide. In the United States (US) alone, there have been 16 million cases and 296 thousand deaths[2]. During the COVID-19 pandemic[3], many politicians have turned to Twitter as a critical media platform to communicate their responses and share public health information to citizens. Twitter is a free microblogging social media platform with over 340 million users and has been a critical platform for world leaders to share information on the spread of the virus, public health recommendations, and government response measures.

Although Twitter can be a powerful tool that political figures can utilize to provide information rapidly and directly, there can often be differences in opinions, and selected facts or fiction that they choose to share. Unlike any global pandemic before, the internet and social media platforms have drastically changed the way we receive and share information. Unlike newspaper articles and journals, tweets can be made within seconds (specifically 280 characters) and are made more frequently (approximately 500 million tweets per day)[4], allowing for a unique perspective in the development of COVID discourse over time (2019-2020). According to Twitter's third quarter results in 2020, there has been a 29%[5] year-over-year growth (a 12% increase from Q3 in 2019). This is the highest growth increase ever recorded for Twitter and this can especially be credited to new improvements to the platform, as well as the dramatic increase in use due to COVID-19 pandemic related discussions. Social media platforms have served as a key role in how individuals are responding to the effects of the pandemic. Political leaders especially have an influence over public opinions when they address public health matters in this manner. While it was found that the majority of politicians' "viral" tweets may be informative about COVID-19, it is important to look at the tweets posted collectively and the terminology that is used within the context of the pandemic (Rufai et al., 2020). As with political figures, individuals have more often posted tweets that are emotionally charged, as well as discuss concerns over public and personal health (Aggarwal et al., 2020). Ambiguity surrounding the nature of COVID, terminology associated with the virus, public health communications, and statements shared by political leaders has led to the cultivation of new words and phrases most frequently used in the context of the pandemic. Keywords such as 'pandemic','corona', 'quarantine', 'mask', 'lockdown' among others are becoming more integrated and frequently used on Twitter discourse in 2020 (Davies, *The Coronavirus Corpora*).

---

[1] Also known as COVID-19, COVID, SARS-CoV-2 (severe acute respiratory syndrome coronavirus 2)
[2] Data provided by Johns Hopkins University Coronavirus Resource Center. Date Accessed December 10, 2020 (See resource Dong,E et.al. )
[3] In this paper, COVID-19 and COVID will be used interchangeably, unless otherwise specified.
[4] Statistic according to https://www.internetlivestats.com/twitter-statistics/ : 6,000 tweets per second, 350,000 tweets per minute. Volume of tweets per year grows around 30%.

      This research aims to investigate COVID-19 discourse on Twitter, specifically that of tweets made by the US politicians. This is further evaluated through the lens of Twitter data from Republican and Democratic political leaders. Through this exploration, we aim to develop an understanding of the terminology used in association to COVID-19 by each political party, most frequent topics addressed, as well as sentiment (left-leaning or right-leaning) of tweets posted. A corpora, created from 12 months of Twitter data, will illuminate the language used by individuals over the course of the pandemic and the differences associated with political parties from this data. Analysis of terminology on the corpora is done using the bag-of-words model, bigram analysis, and TF-IDF. Sentiment analysis, to determine left or right leaning score is done with LinearSVC. The project contributes to current research by providing analysis on language used during the COVID-19 pandemic specific to the US Republican and Democratic politicians, as well as provides insight to ways in which left or right leaning tweets can be predicted based on this new set of terminology.

**2 Related Works**

      Over the course of the pandemic, many research papers in computational linguistics and natural language processing have been published concerning how individuals have responded on social media to the pandemic. Specifically, many studies have looked at Twitter data, as it is frequently updated and is a text rich social media platform which provides powerful insight into responses and use of language over time. Previous studies on Twitter data (Hui Xian Ng et.al 2020), analyzed the emotions expressed on Twitter during the pandemic. Through the use of combined strategies such as topic Hidden Markov Model and sentiment analysis, emotional states of a user were predicted based on the terms used in the Twitter posts (specifically during the COVID-19 pandemic). Ten topics and associated keywords were identified through topic clustering of keywords and a topic identifying modeling tool, Mallet. The topics identified through analysis of tweets made during the pandemic include: general (i.e. news, social, government, fact, question), vaccines/testing (i.e. testing, infection), country specific (i.e. world, spread, nation), lockdown(i.e. citizen, law), call for leader (i.e. time, crisis, action), high case numbers (i.e. death, positive, total), elections (i.e. trump, call, mask), relationships (life, family, friend), concern for healthcare workers (i.e. hospital, health, case), and business and job reliefs (i.e. economy, business).

---

[5] Information provided by :Twitter *Quarterly Results: 2020 Third Quarter*.
https://investor.twitterinc.com/financial-information/quarterly-results/default.aspx (Reported: October 29, 2020)

Although in this research paper, we did not perform sentiment analysis on user emotion, the identified topics and associated keywords served as a guide when analyzing the results of keywords most frequently used by each political party. Analysis of keywords in conjunction allows us to determine topics that may be more frequently used by a political party. Previous work has determined political sentiment using Twitter data, but outside the scope of pandemic (Sujeet & Shetty, 2018). Data related to the UK General Election of 2017 was classified based on the political polarity of the tweets to determine the popularity of the parties using Naïve Bayes (NB), Support Vector Machine (SVM) and k-Nearest Neighbours (kNN). In this paper, the methodology used in sentiment analysis, specifically with NB, SVM will be adapted in our analysis.

Other previous research studies have also analyzed COVID-19 related Tweets from North American (specifically Canada) related to public health communication (Jang, H., et al. 2020). Topic modeling[6] and sentiment analysis are used to identify an individual's opinions on particular topics over time. It was found that key topics Twitter users discussed about public health communications concerning COVID-19 included: social and physical distancing, air travel, handwashing and preventative measures, need to stay home, impact of COVID-19 on essential workers, number of cases and tests, and mask/ face coverings. Frequency of topics based on the number of tweets categorized were analyzed over the time period of January to May 2020. This existing study analyzed the changes in prevalence of particular topics over time. Jang, H. et al. (2020) found that the topic of air travel peaked in February, while the topic of social distancing increased in early March around the time that a nursing home in Seattle, WA had an outbreak of COVID-19. The topic "number of tests and cases" gradually increased over time. This is similar to existing work by Naskar et al. (2020) which found that Twitter users' emotions and opinion on public health issues changed over time.

Additional research explores differences in gender responses to COVID-19, confirming existing gender-linked assumptions (Aggarwal & Stevenson, 2020). From that and many previous studies, (e.g. Mulac et al. (2001); Newman et al. (2008)), it was suggested that men tend towards topics regarding money and occupation whereas women prefer discussion surrounding family and social life. Topics were modeled through the MALLET tool and given coherence scores. Extracting the most prominent topics (from highest coherence scores) allowed for analysis on the categories men and women were preoccupied or concerned with. From the study, both genders, while concerned with general health-issues, had distinctly different topic interests and thereby established the expected gender differences in response to the pandemic. Topic extraction on COVID-19 related data can be used to generate a reliable analysis on the discussions that the general public are attentive to. Hence, our research similarly approaches the question of political differences in exploration of COVID-19 discourse with the intent to discover the important public issues that are important to either party.

Our research is distinct from previous work because it focuses on a specific type of Twitter user (US political leaders) and similarly assesses the topics and frequent keywords used by each political party. Previous works on topic analysis from COVID-19 related Twitter data assists in clarifying topics based on collections of keywords found and defining topic differences between political parties.
This study contributes to ongoing research and lines of inquiry through the following questions:
(1) What terminology and keywords have US politicians acquired when communicating through Twitter during the COVID-19 pandemic?

---

[6] Topic modeling techniques such as Aspect-Based-Sentiment Analysis (ABSA) from which results were interpreted by public health experts. Latent Dirichlet Allocation (LDA) was used for topic modeling to compare the timeline of topic distribution.

(2) Can differences found in terminology and use of language by different political parties during the pandemic help to classify "COVID-19" tweet content as left or right leaning?

## 3 Dataset
### 3.1 Data Sourcing

The dataset used in creating the corpora was collected from Twitter using the *snscrape scraper tool*[7]. The parameters used for the scraper were the date, starting from the first case of COVID, 17th of November, 2019 until 17th of November, 2020, and the usernames of 60 political figures of the US, which sums up to a total of 64,568 tweets. The political figures were chosen as the governors of states and territories of the US, as they were one of the primary actors in addressing the COVID situation, and further the main presidential and vice presidential candidates of 2020 election. The collected data was stored in pandas dataframe and was labelled in each Twitter user's political party, "D" for Democrat, "R" for Republican and "NPP" for New Progressive Party, and the state the user represented.

### 3.2 Data Pre-Processing

Given this primary dataset, the tweets were narrowed down into COVID related themed tweets, in a total of 18,050 tweets, by filtering the tweets to those that contained COVID related terms such as: 'covid', 'covid-19', 'covid19', 'corona', 'coronavirus', 'pandemic', 'sars-cov-2', '2019-ncov', 'virus', 'epidemic', 'flu', 'influenza', 'cold', adapted from terms used by Aggarwal (Aggarwal et al. 2020).

The more focused COVID centered tweets were then tokenized and cleaned in terminology in various ways. Before anything, the first step was to eliminate hyperlinks as they are simply abbreviated urls embedded in the tweets, which we were not interested in studying. Then, the tweets were tokenized by using word_tokenize() function of NLTK library[8]. After the tokenization the corpus was cleaned by eliminating the common stop words, that were obtained from NLTK library combined with other custom words, with the exception of the negation words such as: 'not', 'no', 'n't', so that the corpus could take into consideration of the phrasing context of the tweets. Then, each word was cleaned to the roots into two versions, general stemming and WordNet lemmatization of NLTK library to experiment and to compare results.

### 3.3 Corpora Creation

These corpora were randomly distributed in tweets by shuffling the texts so that each corpus would entail randomized distribution of date range, political figure and political party. Then, the full corpus was separated into development, training and testing corpus, which are divided into 10%, 80%, 10% respectively. The development corpus was used to develop the models, but when training, the development corpus was combined with the training corpus to test the models. Thus, the analysis of the general political sentiments will be done on this combined training data to explain the phenomenon and result obtained from the sentiment analysis implementation to classify the tweets.

## 4 Methodology
### 4.1 Political Sentiments Profiling

To begin understanding the overall dataset at a higher level, we extracted keywords that provide a smaller, more focused set of highly-used words or phrases that can emphasize topics frequently discussed among the political scene on Twitter. Multiple algorithmic approaches were implemented in order to compare and contrast the efficiency of distinctive approaches for our particular use. All the approaches were tested and run on the lemmatized Democratic data, lemmatized Republican data, stemmed Democratic data, and stemmed Republican data.

---

[7] Implemented scraper tools: https://github.com/JustAnotherArchivist/snscrape
[8] https://www.nltk.org/py-modindex.html

### 4.1.1 Bag-of-words Model

Using the bag-of-words model, we can extract the basic features of the text and identify overarching topics under the assumption that a word's higher frequency will ultimately indicate higher importance. A list of distinctive tokens are first generated by saving each word within a dictionary, along with its respective frequency count. By then sorting by frequency (highest to least), the keywords at the top of the list can be indicative of the most influential topics. In comparing the resulting data of the lemmatized files and the stemmed files, top frequencies and keywords were quite similar; however, the lemmatized version produced cleaner results with complete words whereas the stemmed version occasionally only produced the stem counterpart. Therefore, the bag-of-words approach on the lemmatized Republican and Democratic data was prioritized in comparing and analyzing the results.

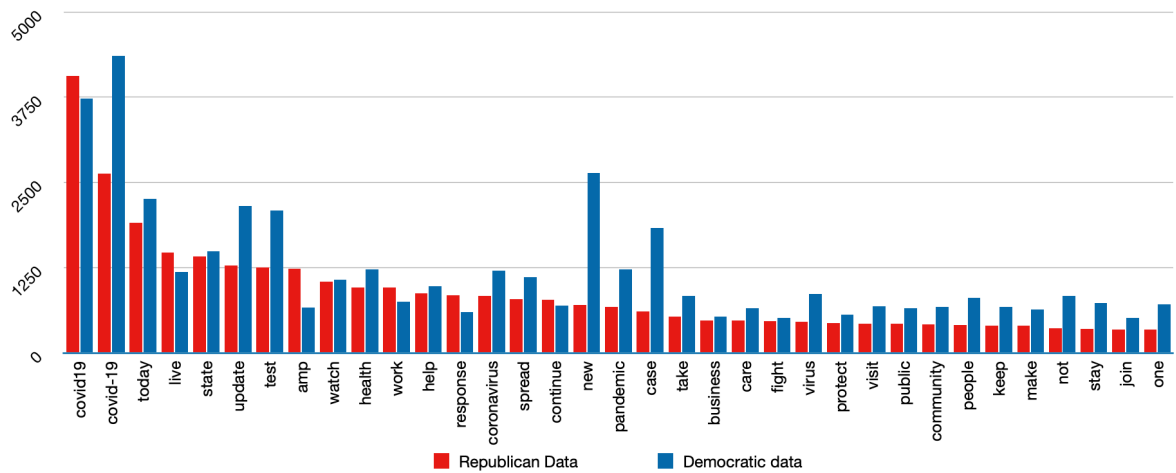

Figure 1: Distribution of Keywords by bag-of-words model. Scaled by frequency.

Analysis of the top sixty keywords were used to develop the visual representations of the bag-of-words model data in Figure 1. The top sixty keywords by frequency were taken from both lemmatized Republican and Democratic datasets. Figure 1 compares the frequency counts of words in the selected portion of the data that appeared in both Democratic and Republican lemma tweet sets. The keywords chosen to be represented in the graph are words that were exactly the same in both sets, allowing for a direct comparison of the frequencies. Spanish keywords that appeared in the bag-of-words model output were removed such that only complete English words were included in the final representation. Stop words, states, and names, were removed from the dataset prior to implementation of the bag-of-words model.

Although the bag-of-words model is a simple approach, it is extremely effective and widely used as a technique for sentiment analysis (El-Din, 2016). Most of the top keywords in either party are neutral or very nuanced words such as "covid-19", "today", "live", etc. However, in comparing the keywords that have a *significant* disparity in frequency between both parties, or in cases where certain words show up in one party and not the other, a more in-depth analysis can be made. Such keywords (known here as "distinct") show much more about the unique characteristics of either party towards COVID-19 and their greater, overall sentiment. The representations for the distinct Republican and Democrat bag-of-words keywords were made in Figures 2 and 3. To further emphasize exactly *how* distinct a keyword is, it was evaluated by a difference count, which is equal to the distinct keyword's frequency in its corresponding party minus the frequency of the same keyword in the other political party.

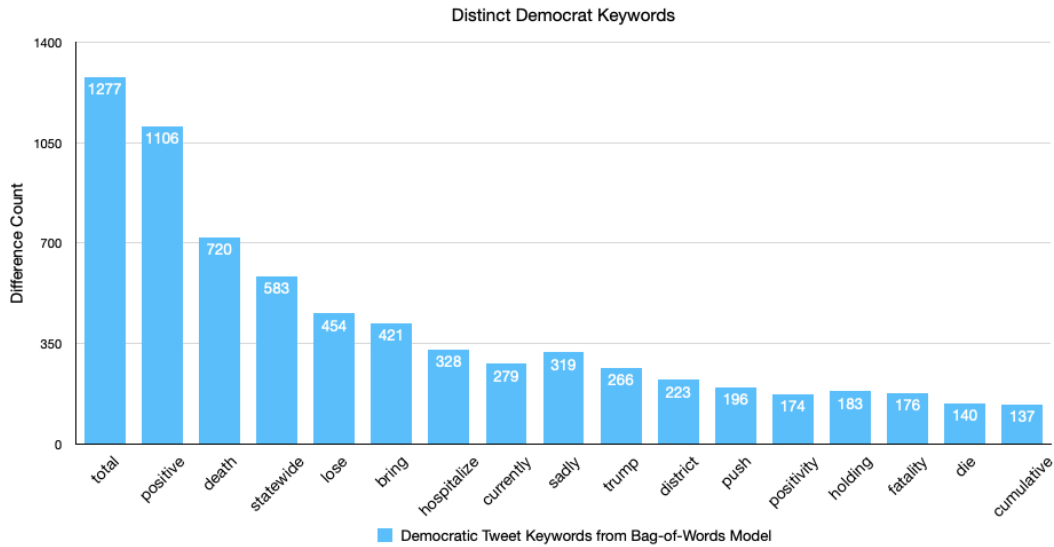

Figure 2: Distinct keywords found considerably more in Democratic party. Count representing frequency difference of keyword in Democratic compared to Republican data.

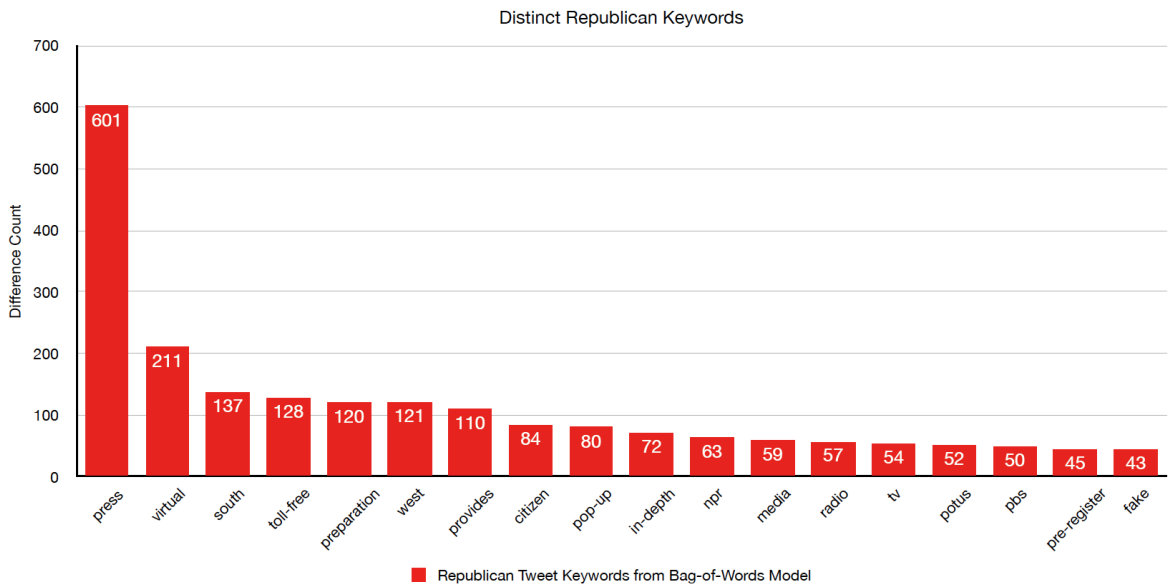

Figure 3: Distinct keywords found considerably more from the Republican party. Count representing frequency difference of keyword in Republican compared to Democratic data.

From the distinct Democratic keywords, there seems to be an overall greater emphasis on the unfavorable aspects of the pandemic, carried by keywords with a more pessimistic tone. For essence, "death" is the third more distinct keyword, followed by "lose", "hospitalize", "sadly" and more. Higher awareness on the casualties of the pandemic seem to be a key feature as well: "total" and "positive" are top Democratic distinct keywords and within the context of pandemic, they are most likely associated with the number of coronavirus cases. Both of these elements may imply a greater sense of solemnity and gravity to the situation, which in turn can explain yet another component in which distinct Democratic keywords mention more coronavirus-related medical precautions or recommendations such as "mask", "home", and "wear." The implications from this data align with conclusions from previous research, that also indicate Democrats' behavior as much more responsive to medical recommendations and in general, Democratic political figures were shown to produce more stay-at-home orders and physical distancing communications then Republicans (Grossman et al. 2020).

      In comparison, the distinct Republican keywords contain a more political attitude towards the coronavirus; the top keywords such as "press", "information", "conference", "briefing", demonstrate more attention to the political responsibilities that the governors may have felt at this time, which involves keeping the public informed about their response, suggestions, and information on the virus. It is possible that Republicans felt more reluctant to accept the severity of the virus, and thereby focused on following its progression throughout the nation. Especially given that beginning from early May, right-leaning media and US President Donald Trump seemed to be outwardly skeptical of the risks from the virus (Grossman et al. 2020).

      Because many of the distinct keywords cannot be analyzed without context on its connotation, only the noticeably strongly-connoted keywords were taken into account. Hence, a bigram model would provide more context to the way in which these keywords are used, and perhaps deeper analysis.

**4.1.2 Bigram Model**

      To better understand the associations between different COVID-19 related keywords, we performed an n-gram analysis on the different political party corpora. We chose to use a bigram model to look more closely at pairs of words such as "social distancing", "wear mask" and "stay home." In other n-gram methods, phrases such as these would not be highlighted. N-gram models approximate the probability of a string of words and help to look more closely at the relationship between words as they appear in a document (in this case, a Twitter post). The bigram model used follows the Markov assumption that the probability of a word depends only on the previous word, such that the bigram probability predicts the occurrence of a word based on the occurrence of the n-1 previous words (Jurafsky and Martin, 2009). Analyzing word pairs of the corpora has provided interesting insight into the frequency of bigram phrases used by each political party selected data. In our particular bigram model, two consecutive words were analyzed at a time. Stop words, punctuation, hashtags, and emoticons were removed from the corpus before being run through the bigram analysis. Bigram count, and frequency were assessed. However based on the output, we felt that the best representation of the data would be by comparing the counts of bigram occurrences. The frequencies and counts of each bigram were then sorted from highest to lowest value and written to a text file. With a bigram model, the phrases and series of words most common in COVID discourse among US politicians in their public Twitter posts could be further compared.

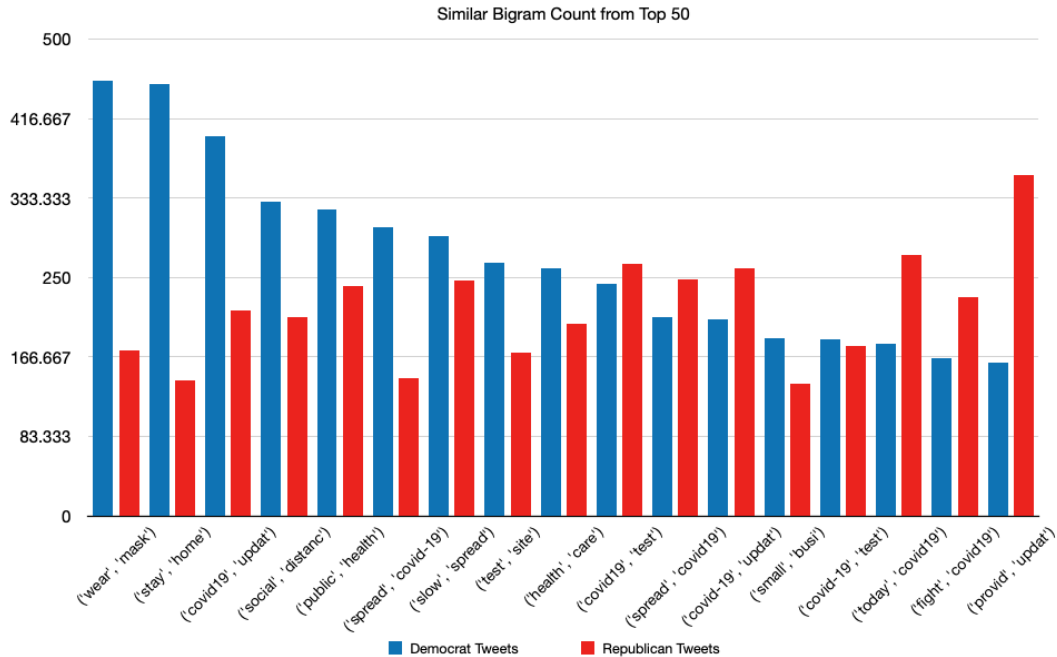

Figure 4: Bigram Phrases by count. Red bars represent occurrence counts of the particular bigrams in Republican tweets. Blue bars represent occurrence counts of particular bigrams in Democrat tweets.

For our analysis, the bigram system was run on separate Democrat and Republican full tweets, stemmed, and lemmatized corpora. However, based on the results, the best outcome was found to be the frequencies and counts of the bigrams based on the stem corpus. This was because there were more word combinations with the root oriented word in the stemmed corporus than the lemmatized corpus. For example in Figure 4, the unigram "distanc" appears in the stemmed form. The bigram "social distanc" can appear in the forms: social distanc*ing* or social distanc*e*. In a corpus data that has been lemmatized, these two bigrams which we consider to have equivalent meanings would be counted as separate bigrams, while in the stemmed corpus, we can better compare the frequency of the overarching bigram we mean to represent. Another example of this that appeared in the output data is the bigram "current hospit" which encompasses the other equivalent bigram counts: "current hospit*alization*," "current hospit*als*," "current hospit*alizations*," among others.

Figure 4 was developed from the top (highest frequency and counts) 50 bigrams determined by our analysis. The input corpora used was the full tweets stemmed corpus for Republican and Democrat datasets. The counts of only exact matching bigram phrases from the highest 50 bigram counts in Republican and Democrat data are shown in Figure 4. From the generated output of the highest fifty bigram counts of separated Democratic and Republican Twitter data corpora, many interesting comparisons can be drawn. In analyzing the data, it was chosen to remove the names of states, people, and hashtags before creating Figure 4 to best represent the bigram count differences for COVID-19 terminology.

The seventeen bigram phrases chosen for Figure 4 are drawn from comparing the exact matching bigram phrases that appeared in the top fifty bigram counts in both Republican and Democrat data. Although Figure 4 provides a small portion of the larger dataset, it is helpful to distinguish these comparisons and inspect the interesting outcomes. For example, it can be seen that the bigram phrase ('wear','mask') appears in the Democratic tweets 456 times, while in the Republican tweets, the bigram only appears 174 times. The phrase "wear mask" appears 282 more times in the full democratic data than in the Republican data. Other terms relating to public health actions such as ('stay','home') and ('social', 'distance') are also seen to appear significantly more times in the Democrat tweets. On the other hand, in the Republican tweets, counts for bigrams that have to do with "updating" on COVID-19 related issues appeared more often in the Republican tweet set. As with the bigram phrases ('provid', 'updat') which has a count of 357 in the Republican data and 161 times in the Democrat data. However, as can be seen in Figure 4, the majority of bigrams are used with similar frequencies throughout the Twitter corpora. This supports our claim that a unique set of terminology and language surrounding discussions about COVID-19 exists and extends across political parties. Given the matching bigrams that appear with similar frequency such as ('public', 'health'), ('slow', 'spread') , ('covid19','test'), and ('small', 'busi') , it can be seen that similar discussions and contexts relating to public health matters concerns, and goals, as well as testing and local economies appear in Twitter posts shared by both political parties.

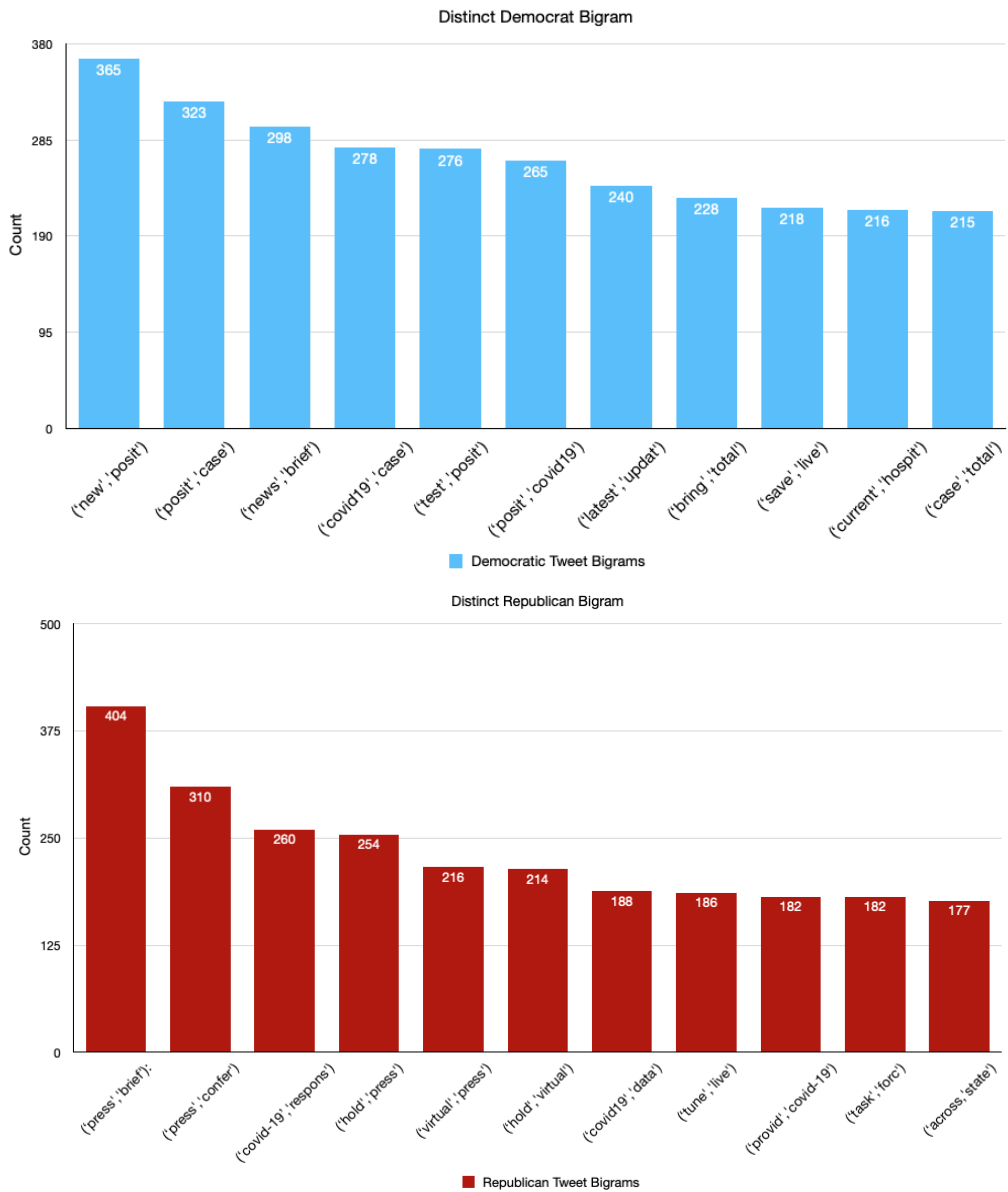

Figure 5: Distinct Political Party Bigram phrases from portion of fifty most frequent bigram phrases (excluding matching bigrams)

To further this analysis, a portion of the list comparing the unique Democrat and Republican bigram phrases from the highest fifty counts is represented in Figure 5. Any matching bigram phrases were removed to better understand the differences of the top unique phrases used by each political party. It is interesting to note that the most frequent bigrams from the Republican corpus that did not appear in the highest fifty counts of the Democrat corpora are of a similar topic, and the same is for the most frequent unique Democrat bigrams. The most frequent unique Republican bigrams surround the topic of "press" and updating the public: ('press', 'brief'), ('press', 'confer'), ('hold', 'press'), ('virtual', 'press'), and ('tune', 'live'). Although this is an interesting comparison to draw, these bigrams most likely appear more frequently due to tweets and retweets made concerning the US COVID19 task force and made by the Republican president Donald Trump and other state government briefings (Yaqub, Ussama. (2020)). The topics of press coverage, task force, and updates to COVID19 case data appear to encompass the categories associated with the top bigrams. In the Democrat data, the topics that can be drawn are concerning updates on positivity rates, hospitalizations, and similarly COVID-19 related data update. The top unique bigrams all concern discussions of positivity rates and ('latest', 'updat') - latest update(s) to the growing numbers of pandemic. This topic[9] was also previously identified in work by Hui Xian Ng et. al (2020). Similarly the topic of "numbers of tests and cases" was also found as a distinct COVID-19 topic[10] based on analysis by Jang, H. et al (2020). This confirms that the topics generalized based on this bigram keyword analysis are inline with previous research topic identification for COVID-19 related Tweets.

As with the matching bigrams, it is again recognized that the majority of found COVID-19 related bigrams from this Twitter data indicate that similar phrases are used by both political parties and that similar topics are discussed. While differentiating the unique bigrams of each party, highlight the differences in frequent topics that appear in Twitter posts. This understanding helps to better classify a left or right leaning tweet based on the frequency of particular bigram phrases from each political party. The analysis of the corpora with a bigram model helps to better understand further tweet classifications made in sentiment analysis and will help evaluate any errors in classifying a tweet.

**4.1.3 TF-IDF Model**

The bag-of-words model allowed researchers to see what words were used in one political party's tweets and analyze overall frequencies by putting equal weight on each word. However, knowing that not every word is emphasized evenly by each political party and even each politician. A closer study on words used in each tweet of one politician and contrasting it to the tweets of the other party is needed to reveal the weighted emphasis in a timeline manner, so TF-IDF methodology was chosen for this purpose.

Originally, the time-unordered dataset was used for the TF-IDF test. Target keyword is the next word in the tweet of the current political party, while the dataset is the entire tweets from the compared political party. Specifically, for each word in one tweet of a party, its rate of exposure in the tweet that it belongs to was first calculated (TF), then the same word's rate of exposure among all tweets of the other political party was computed (IDF). From there, each word's TF-IDF value, which represents how much a word is more emphasized in its document than in the dataset, was obtained in a given tweet (TF-IDF). Then, a for loop was used to filter out the word with the maximum TF-IDF value in that tweet. Therefore, the resulting output file contains words with maximum TF-IDF value in each tweet of the comparing political party.

---

[9] Top ten topics and sample keywords identified in Hui Xian Ng et. al (2020) : High Case Number (death, case, positive, total, high)

[10] Representative keywords associated with the topic identified as "number of tests and cases - data specific" are : positive, test, cases, patients, hospital, data. These keywords are similar to those found in both political party bigram phrases.

Unfortunately, the first tryout failed due to the scattered, untelling word pool generated: no apparent repetition of words was observed and the TF-IDF value of each word was relatively low. Reflecting on the insignificance of the result, two things came to mind. First, treating the entire tweet history of one political party as the dataset may be an important factor that dilutes the emphasis of a word. Since tweets starting in November of 2019 till November of 2020 were scraped for use, this large amount of data encapsulated a great variety of context. Therefore, for time-sensitive events that appeared at different time and date, it may be inappropriate to compare its emphasis in its context to the entire tweet dataset which includes too much variability. Second, Twitter, as a news and opinion delivery platform implies time effectiveness. Comparing the usage of a word to that of a year ago may lose the context relevance, so the second modification needs to express time and context relevance.

Together, re-populated data sorted by time and date was used for new computation. Moreover, instead of comparing each word in a tweet to another tweet dataset, every ten tweets from the other political party were used as the dataset for comparison, because it was observed that there were about ten tweets from each day of a political party provided by the data. Therefore, the time frame is restricted to one day to deepen the study scope. Nested while and for loops were used so that each tweet in the first ten tweets from one party were compared to ten tweets of the other party. As expected, the result yielded a much more significant relationship among words and higher TF-IDF values. The top 20 most repeated max TF-IDF words for each party were filtered out for a closer analysis. In Figure 6, the orange bars represent the most frequently appeared word that has the highest TF-IDF value in its tweet of the Republican while blue bars represent that of the Democratic party. The keywords are ordered from most frequent to least frequent, and the specific word as well as its repeated quantity is marked above the bar it represents. Both Republican and Democratic parties had keywords related to COVID-19 as the most frequently repeated, with the word 'covid19' used by Republican politicians for 48 times and 'covid' by Democratic politicians for 38 times. Rest of the top 20 most repeated max TF-IDF words are also mostly related to COVID-19, topics including attitudes, policies and emotions.

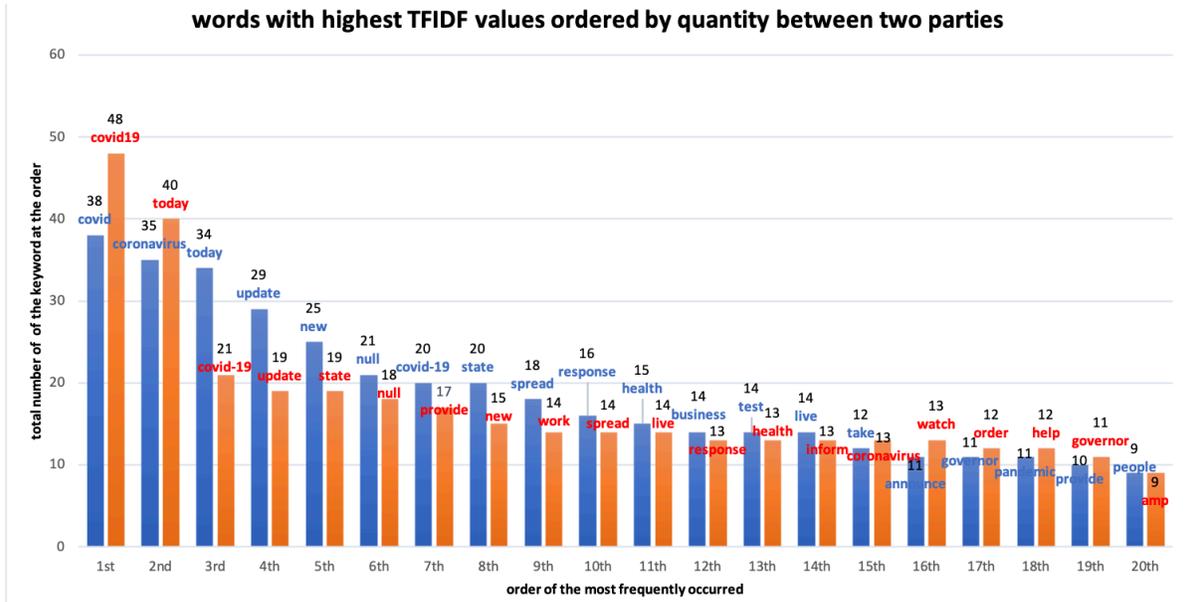

Figure 6: top TF-IDF values for keywords from Democratic (blue) and Republican (orange) data

To further view the category of the words, Figure 7 and Figure 8 colored each word so that the same category has the same color. The circle on the left represents words used by the Democratic party while the circle on the right represents words used by the Republican party.
- Words related to **the virus** are colored in bright pink;
- Words related to **people and lives** are colored in bright yellow;
- Words related to **policy actions** are colored in bright blue;
- Words related to **the government** are colored in light purple;
- Words related to **wellness support** are colored in light pink;
- Words related to **business** or work are colored in orange;
- Words related to **emotions** are colored in green;
- Word **'test'** is colored in dark purple; word **'stay'** is colored in light blue;
- Word **'spread'** is colored in dark green;
- Other words that do not have a close relationship to the COVID-19 event are left in white.

Figure 7 and Figure 8 record the first 20 max TF-IDF words used by two parties chronologically, to further see the categories of topics in a timeline manner. By result, we observe more empathetic motion words were used by the Democratic party, the direct emphasis on the virus was mentioned earlier by the Democratic party than by the Republican party, and more action or policy related words were mentioned by the Democratic party.

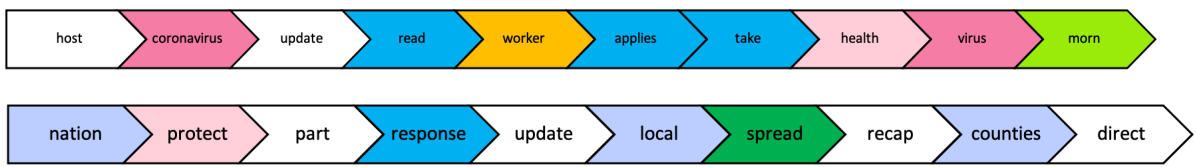

Figure 7: first 10 max TF-IDF words chronologically
Democratic(top), Republican(bottom)

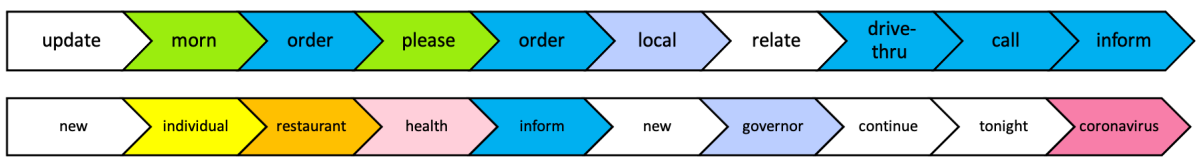

Figure 8: 10th to 20th max TF-IDF words chronologically
Democratic(top), Republican(bottom)

Further, to discuss some key observations from the appearance of the categories in each party: the only emotion words appearing in the top 20 most repeated max TF-IDF groups were from the Republican party, with the words "please" and "thanks." Additionally, both parties mentioned topics around the virus with similar rates of exposure. The word "stay" only appeared in the Democratic top 20 most repeated max TF-IDF group while word 'test' appeared in the Republican's.

However, because each tweet was processed and computed chronologically in the TF-IDF model, two parties' emphasis and attitudes at a point of time were captured by the max TF-IDF words which bag-of-words and bigram model could not achieve. From the result, the Democratic party raised the attention of COVID-19 earlier than the Republican party and included more policy or action related announcements than the Republican. To see more patterns or shifts of each party's emphasis, more max TF-IDF words down the timeline should have been analyzed. Also, for the occurrences of the less relevant keywords appearing in the max TF-IDF results, one possible reason may be the assumption that each day contains about 10 tweets by each party is not accurate enough, therefore causing unwanted noise.

**4.1.4 Results and Discussion**

We compared keyword trends in Republican data and Democratic data for each model. To present an overview of the results from all of the approaches altogether, the top distinctive keywords and phrases were collected for the figure below.

| Model | Top Distinctly Democratic | Frequency Difference | Top Distinctly Republican | Frequency Difference |
|---|---|---|---|---|
| bag-of-words | "total" | 1277 | "press" | 601 |
| | "positive" | 1106 | "virtual" | 211 |
| | "death" | 720 | "hutchinson" | 149 |
| | "statewide" | 583 | "south" | 137 |
| | "lose" | 454 | "realdonaldtrump" | 125 |
| | "bring" | 421 | "preparation" | 120 |
| | "hospitalize" | 328 | "toll-free" | 128 |
| | "currently" | 279 | "west" | 121 |
| | "sadly" | 319 | "provides" | 110 |
| | "trump" | 266 | "citizen" | 84 |
| bigram | "('wear', 'mask')" | 282 | "('press', 'brief')" | 352 |
| | "('stay', 'home')" | 311 | "('provid', 'update')" | 196 |
| | "('new', 'posit')" | 359 | "('press', 'confer')" | 260 |
| | "('posit', 'case')" | 282 | "('covid-19', 'response')" | 146 |
| | "('covid-19', 'case')" | 226 | "('hold', 'press')" | 232 |
| | "('posit', 'case')" | 282 | "('virtual', 'press')" | 213 |
| | "('updat', 'new')" | 304 | "('hold', 'virtual')" | 213 |
| | "('updat', 'covid-19')" | 231 | "('covid19', 'data')" | 184 |
| | "('spread', 'covid-19')" | 159 | "('tune', 'live')" | 161 |
| | "('news', 'brief')" | 289 | "('provid', 'covid-19')" | 180 |
| TF-IDF | "covid" | 38 | "covid19" | 48 |
| | "coronavirus" | 22 | "today" | 6 |
| | "update" | 10 | "provide" | 7 |
| | "new" | 10 | "work" | 14 |
| | "business" | 14 | "inform" | 13 |
| | "test" | 14 | "watch" | 13 |
| | "take" | 12 | "order" | 12 |
| | "announce" | 11 | "help" | 12 |
| | "pandemic" | 11 | "amp" | 9 |
| | "people" | 9 | | |

Figure 9: Collection of top Democratic and Republican words/phrases from each model. Bag-of-words and bigram evaluated by frequency count and TF-IDF evaluated by TF-IDF scores.

For Figure 9, a Python program was written to find distinct keywords in each party through each model's output. A keyword is considered "distinctly" one party if it appears significantly more frequently in that party than the other, or if it only exists in that party. For the bag-of-words model, because frequencies ranged from as little as 1 all the way to 4353, a keyword is distinct if it was at least five times more frequent in one party than the other. In comparison, the bigram frequencies had a much, much smaller range with the highest count being 354. For that reason, following the same rule of a frequency five times greater would produce very little and unreasonable results. Instead, for bigram, a keyword is considered distinct if it was at least two times more frequent than in the other party. Even more than the bigram model, the TF-IDF values were far smaller and closer together. Thus, when the TF-IDF keywords were also run through the same program, a "distinct" keyword appeared if it had at least a score of 5 TF-IDF values higher than in the other party. Thus, the TF-IDF value difference of keywords between the two political parties is represented under Frequency Difference. It is important to note, however, that the TF-IDF values are different from the usual direct measures of frequency as evaluated by the other two models. In an effort to show more concretely and mathematically how distinct the keyword in one party to the other, the difference was calculated by subtracting the keyword's frequency in the other political party from the keyword's frequency in the focused, distinct political party. The results from the program alone produced many "unratable" keywords that could not present any reliable information. Two annotators filtered out keywords from the results through the following criteria: keyword was an area or state name, keyword was in another language besides English, keyword was not a valid word, or keyword was an unclear or unintelligible acronym. Keywords that contained names were left alone. Only the keywords established by both coders as rateable were included in the table for analysis.

From the culmination of all three models, we can conclude several distinctions in relevant and important topics between both political parties by evaluating the top distinctly Democratic and Republican keywords and phrases. All three models seemed to reach a consensus on a few, key pivotal partisan differences. In general, the Democratic Party was far more concerned with the casualties of and data surrounding the pandemic, and emphasized related keywords quite frequently, such as "total", "positive", "death", "('posit', 'case')", "test." Additionally, Democrats were found to be more action-oriented with keywords related to giving medical precautions or recommendations to the public (e.g. "hospitalize", "('wear', 'mask')", "('stay', 'home')", "test"). In contrast, the Republican party seemed to be more invested with their political responsibilities in updating the public and keeping watch on the progress of the virus; distinct keywords related to this include "press", "virtual", "('press', 'brief')", "('provid', 'update')", "today", "inform."

Further nuances of partisan differences were revealed by different model approaches individually. From just the bag-of-words model alone, it was clear that the Democratic data used more negatively-connotated words that ultimately brought greater gravity to the pandemic. The bigram model provided further connotation to many of the keywords found in the bag-of-words model, and established the influential policy-oriented phrases that arose due to the pandemic. While the TF-IDF model appeared to have the least disparity in distinct keywords across the three approaches, the model suggested that many topics were actually mentioned with similar rates of exposure. In particular, this is supported by the bigram model, which also found that both parties were invested in discussion surrounding testing and the pandemic's effect on the economy.

### 4.2 Political Party Alignment Classification

After understanding the overall usage and frequency of usage of words and phrases used by the different political parties, using various techniques listed above, we performed a sentiment analysis on each tweet to classify a tweet based on the user's affiliation to a political party.

### 4.2.1 Political Party Alignment Labeling

After understanding the overall usage and frequency of words and phrases used by the different political parties, using various techniques listed above, we performed a sentiment analysis on each tweet to classify a tweet based on the user's affiliation to a political party. For the scope of the paper, we looked specifically at the two sentiments, the left and the right, grouping the Democrat and the New Progressive Party as the left and the Republican Party as the right, and expressed the left with label of "1" and the right with a label of "-1", for the simplicity of our analysis.

| Left-Leaning Tweets (+1) | 8,979 | 55.3% |
| --- | --- | --- |
| Right-Leaning Tweets (-1) | 7,269 | 44.7% |
| Total Tweets | 16,248 | 100% |

Figure 10: Class Distribution

### 4.2.2 Tweets Classification

In this paper, we considered two classification algorithms: SVM and NB classifiers, based on Sharma and Shetty (2018). Specifically, we implemented a Multinomial NB Classifier (MultiNB)[11] and Linear SVM Classifier (LinearSVC)[12], given the works that have been done in sentiment analysis before.

### 5 Experiment Results

As we got interesting results in exploring the language models (bag-of-words and bigram) and TF-IDF modeling with the corpora, we applied these different methods to extract features for the previously mentioned classification methods. In vectorizing each tweet, we experimented with both the normally used CountVectorizer[13] and TfidfVectorizer[14] to fully apply our understanding of normalization of frequency of the words that we have attained on TF-IDF analysis. We built classifiers on two language models: bag-of-words and bigram on both stemmed and lemmatized corpus and compared their performance in predicting the correct political party alignment.

When running the classification algorithm in different models, the results were as follows: Figure 11 for CountVectorizer and Figure 12 for TfidfVectorizer. By analyzing the experimental results as shown in both tables, the combination of LinearSVC classification algorithm and lemmatized corpora had the accuracy. This result aligns with the related works done by Sharma and Shetty (2018), with LinearSVC slightly outperforming MultiNB. Also, as the combination of LinearSVC classification algorithm and stemmed corpora generally yields a higher accuracy than of LinearSVC classification algorithm and lemmatized corpora, which makes it plausible that the classification algorithm had a higher impact on the accuracy than the cleaning model.

---

[11] Scikit-learn Module: Naive Bayes Multinomial NB
https://scikit-learn.org/stable/modules/generated/sklearn.naive_bayes.MultinomialNB.html
[12] Scikit-learn Module: Linear SVC
https://scikit-learn.org/stable/modules/generated/sklearn.svm.LinearSVC.html
[13] Scikit-learn Module: Feature Extraction - Count Vectorizer
https://scikit-learn.org/stable/modules/generated/sklearn.feature_extraction.text.CountVectorizer.html
[14] Scikit-learn Module: Feature Extraction - TFIDF Vectorizer
https://scikit-learn.org/stable/modules/generated/sklearn.feature_extraction.text.TfidfVectorizer.html

Furthermore, comparing between the two vectorization methods, we can observe that while the model with features using bigram performed better with CountVectorizer method, the model with features using bag-of-words performed better with TfidfVectorizer method, with accuracy 0.927 compared to 0.922. Overall, the experimented results show that for the bag-of-words model, the accuracies of CountVectorizer in bag-of-words is around 0.010 less than the accuracies of TF-IDF in LinearSVC model, however in MultiNB, CountVectorizer outperforms with 0.001. However, in the bigram model, there is a larger difference in accuracies between the two vectorization, with CountVectorizer outperforming by around 0.060 for both LinearSVC and MultiNB.

|  | Bag-of-Words | | Bigram | |
| --- | --- | --- | --- | --- |
|  | Stem | Lemma | Stem | Lemma |
| # of features | 12,440 | 14,590 | 131,846 | 137,714 |
| Accuracy - LinearSVC | 0.908 | 0.911 | 0.921 | **0.927** |
| Accuracy - MultiNB | 0.892 | 0.898 | 0.913 | 0.912 |

Figure 11. Classification Accuracy of using LinearSVC and NB models of Different n-gram Implementations - CountVectorizer

|  | Bag-of-Words | | Bigram | |
| --- | --- | --- | --- | --- |
|  | Stem | Lemma | Stem | Lemma |
| # of features | 12,440 | 14,590 | 119,406 | 123,124 |
| Accuracy - LinearSVC | 0.918 | **0.922** | 0.861 | 0.867 |
| Accuracy - MultiNB | 0.891 | 0.891 | 0.849 | 0.853 |

Figure 12. Classification Accuracy of using LinearSVC and NB models of Different n-gram Implementations - TfidfVectorizer

## 6 Evaluation

The discrepancy in performance in CountVectorizer and TfidfVectorizer of the experimented result aligns with our understanding of the language models and vectorization methods. With a bigram, there will be significantly more features that the model will be trained upon. This in CountVectorizer would work in favor of getting a higher accuracy, as it would have more features to count and weigh in testing the model and predicting the political party. However, in TfidfVectorizer, compared to a bag-of-words model, a feature in a bigram model would not be given as much emphasis, as the increase of number of features means scattered attention given for each feature, when normalized. Thus, it would have more focus in forming a more general feature in a tweet, which would lose accuracy and precision in making a prediction. Therefore, from the build of the model itself, the bag-of-words model will yield a sharper model in TF-IDF vectorization while, bigram model will weaken the model by adding distracting features, contributing to added misclassified/ambiguous tweets.

Further, when running more analysis on the measures of the model, we see that the left-wing tweets are generally better performing than the right, which is consistent with the fact there were more Democratic tweets in the corpora as mentioned in Political Party Alignment Labeling of the combined training corpus of 55.3% left and 44.7% right. Thus, a limitation in the sentiment analysis could simply come from this distribution difference.

|           |              | Class | Precision | Recall | F-Measure |
|-----------|--------------|-------|-----------|--------|-----------|
| LinearSVC | Bag-of-Words | +1    | 0.915     | 0.924  | 0.920     |
|           |              | -1    | 0.905     | 0.895  | 0.900     |
|           | Bigram       | +1    | 0.929     | 0.940  | 0.934     |
|           |              | -1    | 0.924     | 0.911  | 0.918     |

Figure 13. Additional measure of LinearSVC on Different n-gram Implementations - CountVectorizer

|           |              | Class | Precision | Recall | F-Measure |
|-----------|--------------|-------|-----------|--------|-----------|
| LinearSVC | Bag-of-Words | +1    | 0.923     | 0.937  | 0.930     |
|           |              | -1    | 0.920     | 0.903  | 0.912     |
|           | Bigram       | +1    | 0.868     | 0.896  | 0.881     |
|           |              | -1    | 0.866     | 0.831  | 0.848     |

Figure 14. Additional measure of LinearSVC on Different n-gram Implementations - TfidfVectorizer

**6.1 Error analysis**

As determined by bag-of-words, bigram, and TF-IDF analysis, a large majority of COVID related keywords appear in both Republican and Democratic datasets at similar rates. Common misclassifications occur in features that appear more frequently in one political party than another. Since the majority of keywords appear in both political party data, but at different prevalence levels or "distinct" values, the most common form of misclassification appears to be based on the mis-assignment of a tweet based on keyword frequency within a political party. Additionally, some misclassifications come from the model not being able to fully grasp the nuances and the context in which the features have been used.

We analyze some examples of misclassified tweets from all models in Figure 15 to derive plausible reasons behind each misclassification.

| Tweets (cleaned and tokenized) | Prediction | Actual | Reason for Misclassification |
|---|---|---|---|
| slow spread COVID19 close pre-K K-12 school high ed insts begin casino racetrack theater gyms non-essential retail recreational amp entertainment biz gathering | -1 | 1 | List of venues like "casino", "racetrack", "theater", "gyms", "retail", "recreational", "entertainment", "biz", "gathering" and their combination had higher mentions in Republican tweets than Democrat tweets. Also, the frequency of ('close', 'pre-K') was first mentioned in the test corpus, and not trained. Thus, although the connotation of the tweet with the word "close" gives a very different insight of the tweet, the model was not able to grasp the nuance of the sentence as accurately. |
| Testing critically important move forward fight COVID19 continue work closely hospital university health system commercial lab physician across state increase capacity deploy supply need | -1 | 1 | ('across','state') is found in the distinct Republican bigram. It appears in a count of 177. This bigram does not appear in the highest 50 most frequent bigrams for the Democratic data. (see Distinct Republican bigram in Figure 5). Additionally ('fight', 'covid') appeared more frequently in the bigram for Republican tweets than Democratic tweets ( See Figure 4). Both of these keyword phrases could have caused misclassification. The keyword 'test' is also more frequent in the bag-of-words Republican data. |
| beginning pandemic July half hospitalization come large urban county see people rural area hospitalize COVID19 | 1 | -1 | Lemma bag-of-words model keyword 'hospital' appeared in the Republican set with a frequency count of 96 and in the Democratic set with a frequency count of 267. Term 'hospitalize' appeared most frequently in Democratic data (based on bag-of-words). For bigram analysis, 'hospital' unigram stem appears in top distinct dem data, but most frequently in the bigram ('current','hospit') rather than ('hospit', 'COVID-19'). Since this word appeared more frequently in the Democrat sets of data, the terms ranking may have caused misclassification. |
| Join LIVE discuss late update regard COVID19 | 1 | -1 | While 'LIVE' appeared more frequently in the Republican bag-of-words data, the keyword 'update' appeared even more frequently in the democartic bag-of-words output. However, bigram comparison shows that bigram ('provid', 'update') appeared more frequently in Republican tweets than Democratic tweets. Since the unigram was not used in this particular bigram order in this tweet, possible reason for misclassification. |

Figure 15. Examples of Misclassification.

**7 Conclusion and Future Work**

As previously discussed, results from all three models (the bag-of-words, bigram, and TF-IDF), all suggest that Democrats contain more keywords and phrases related to the casualties of the pandemic, and tend to give more action-oriented medical recommendations. On the other hand, the Republicans used more keywords that are associated with their political responsibilities, such as keeping the public updated throughout media conferences and carefully watching the progress of the virus. In implementing a systematic approach to predict and distinguish a tweet's political stance (left or right leaning) based on its COVID-19 related terms, we found the SVM classification and lemmatized corpora in general to be better performing than their comparisons. Also, depending on their vectorization methods, we observed different performance for different language models. The CountVectorizer on the bigram model produced the highest accuracy of 0.927 throughout the different comparisons and the TfidfVectorizer on the bag-of-words model produced an accuracy of 0.922.

There is much future work that can be done using this research model and analysis. First, the paper's model and analysis could be improved by making systematic improvements. An improvement we could make would be to improve on the current model and the implementation of the sentiment analysis and our understanding of US political discussion on COVID-19 pandemic would be to increase the quantity and the quality of the data. This could be done by simply increasing the corpora by scraping tweets from all political representatives, like senators, or being more purposeful and experimenting in how the cleaning of the data and removing stop words would impact the accuracy. Additionally, apart from the n-gram models explored in the paper and the classification algorithms explored in the paper, more could be done.

Further work that could be done would be to extend the models of this paper to better understand the terminology and keywords used by global political leaders on Twitter during the COVID-19 pandemic. The corpora and data could be expanded to include political leaders of different countries and comparisons could be drawn between keywords discovered by country rather than by party. Also, sentiment analysis could also be enhanced to assign a tweet to the country it may be generated from based on the nature of the terminology. It is likely that, as with comparing political parties, the terminology related to COVID-19 and topics/keywords discovered may be similar across the world. This work would be interesting to implement to note nuances in the different keywords and topics discussed on Twitter by world political leaders. Moreover, it could be used to investigate the policies and positions of each country with the language usage with predicting the COVID-19 trends and seeing the relationship between them.

**Acknowledgements and Work Distribution**

This project was created in collaboration through New York University's NLP (Natural Language Processing) course.

We would like to thank Professor Adam Meyers for his support, suggestions, and assistance with the final project. Thank you for an amazing virtual semester in Natural Language Processing!

| | |
|---|---|
| Cindy Kim | Abstract, Related Works, Bag-of-words Model, Political Sentiments Profiling - Results and Discussion, Sentiment Analysis - Conclusion and Future Work |
| Daniela Puchall | Introduction, Related Works, Bag-of-words graphs, Bigram Model, Sentiment Analysis - Error Analysis, Conclusion and Future Work |
| Jiangyi Liang | TF-IDF Model |
| Jiwon Kim | Related Works, Data Sourcing/Cleaning, Sentiment Analysis - Classification/Labeling, Experimented Results, Evaluation, Conclusion and Future Work |